\pdfoutput=1

\documentclass[11pt]{article}

\usepackage[preprint]{acl}

\usepackage{times}
\usepackage{latexsym}
\usepackage{graphicx}
\usepackage{subcaption}
\usepackage{multirow}
\usepackage{float}
\usepackage{amsmath}
\usepackage{enumitem}
\usepackage{microtype}
\usepackage{booktabs} 

\usepackage[ruled,vlined,noend]{algorithm2e}
\usepackage[T1]{fontenc}
\usepackage{lipsum}
\usepackage{tcolorbox}
\usepackage{mdframed}
\usepackage[utf8]{inputenc}

\usepackage{microtype}

\usepackage{inconsolata}

\usepackage{graphicx}

\newcommand{\redcross}{%
    \includegraphics[height=1em]{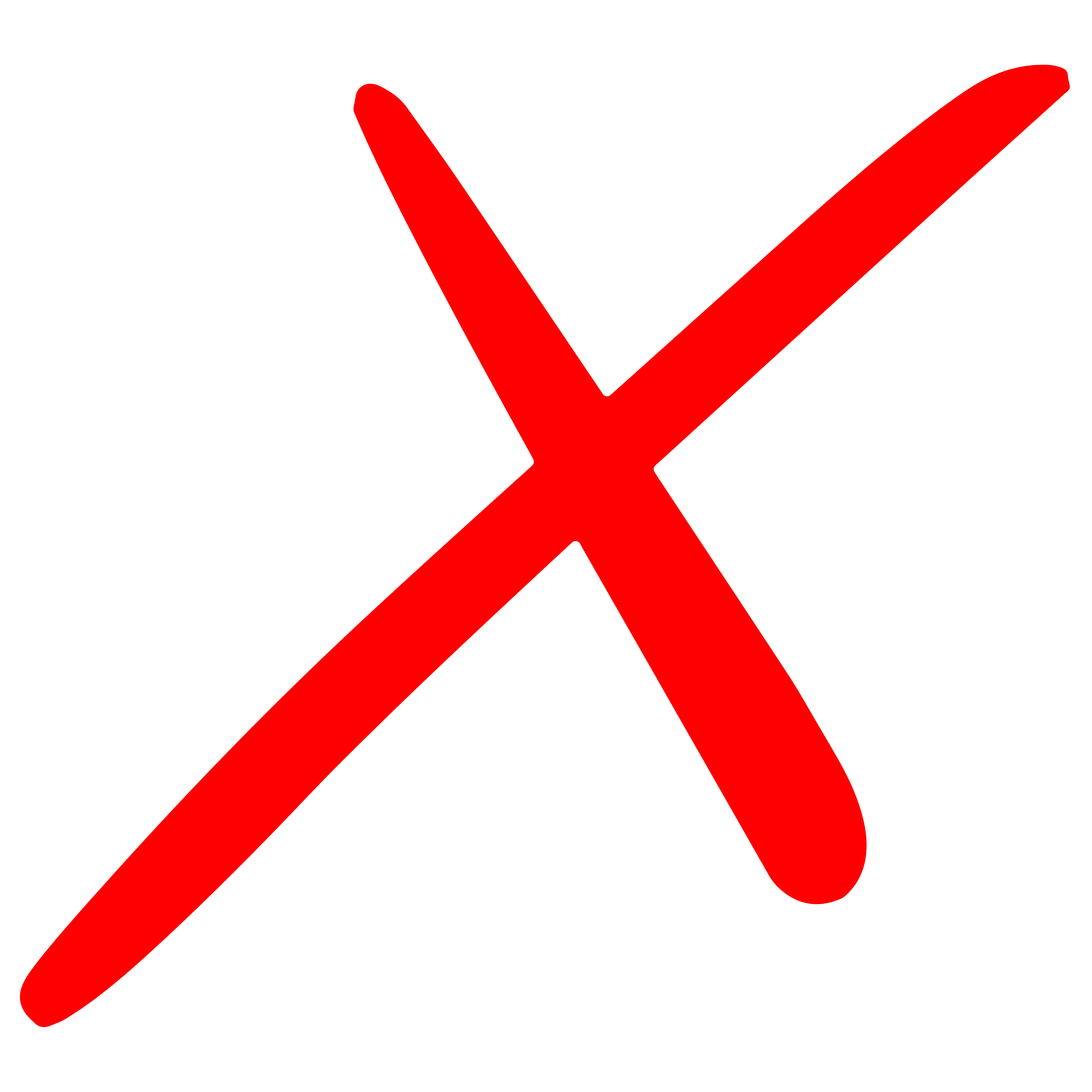}%
}
\newcommand{\greencheck}{%
    \includegraphics[height=1em]{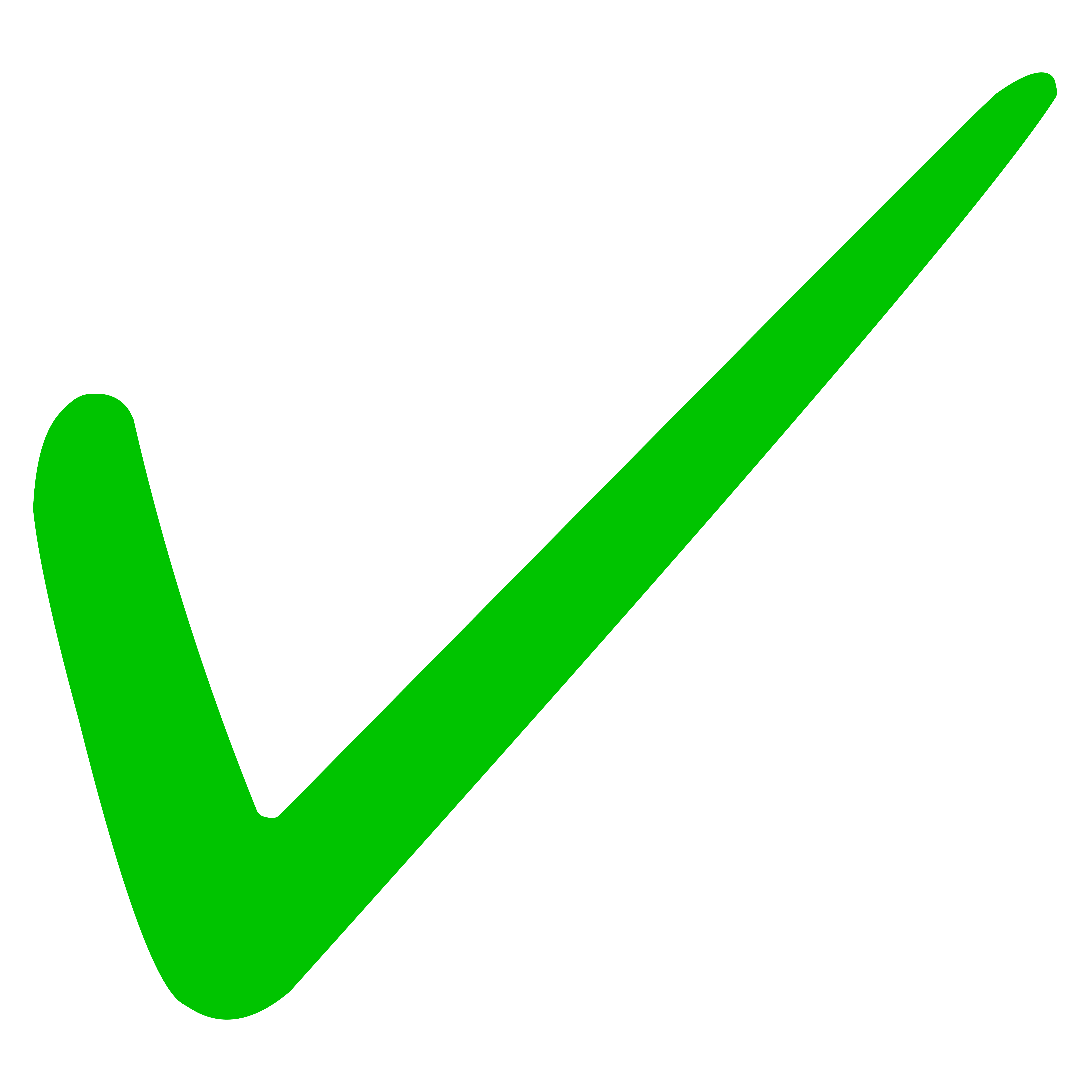}%
}

\title{ExpertGenQA: Open-ended QA generation in Specialized Domains}

\author{
 \textbf{Haz Sameen Shahgir\textsuperscript{1}},
 \textbf{Chansong Lim\textsuperscript{1}},
 \textbf{Jia Chen\textsuperscript{1}},
 \textbf{Evangelos E. Papalexakis\textsuperscript{1}},
 \textbf{Yue Dong\textsuperscript{1}},
\\
\\
 \textsuperscript{1}University of California Riverside
}

\begin{document}
\maketitle

\begin{abstract}
Generating high-quality question-answer pairs for specialized technical domains remains challenging, with existing approaches facing a tradeoff between leveraging expert examples and achieving topical diversity. We present ExpertGenQA, a protocol that combines few-shot learning with structured topic and style categorization to generate comprehensive domain-specific QA pairs. Using U.S. Federal Railroad Administration documents as a test bed, we demonstrate that ExpertGenQA achieves twice the efficiency of baseline few-shot approaches while maintaining $94.4\%$ topic coverage. Through systematic evaluation, we show that current LLM-based judges and reward models exhibit strong bias toward superficial writing styles rather than content quality. Our analysis using Bloom's Taxonomy reveals that ExpertGenQA better preserves the cognitive complexity distribution of expert-written questions compared to template-based approaches. When used to train retrieval models, our generated queries improve top-1 accuracy by $13.02\%$ over baseline performance, demonstrating their effectiveness for downstream applications in technical domains.
\end{abstract}
\section{Introduction}
\begin{figure*}
    \centering
    \includegraphics[width=1\linewidth]{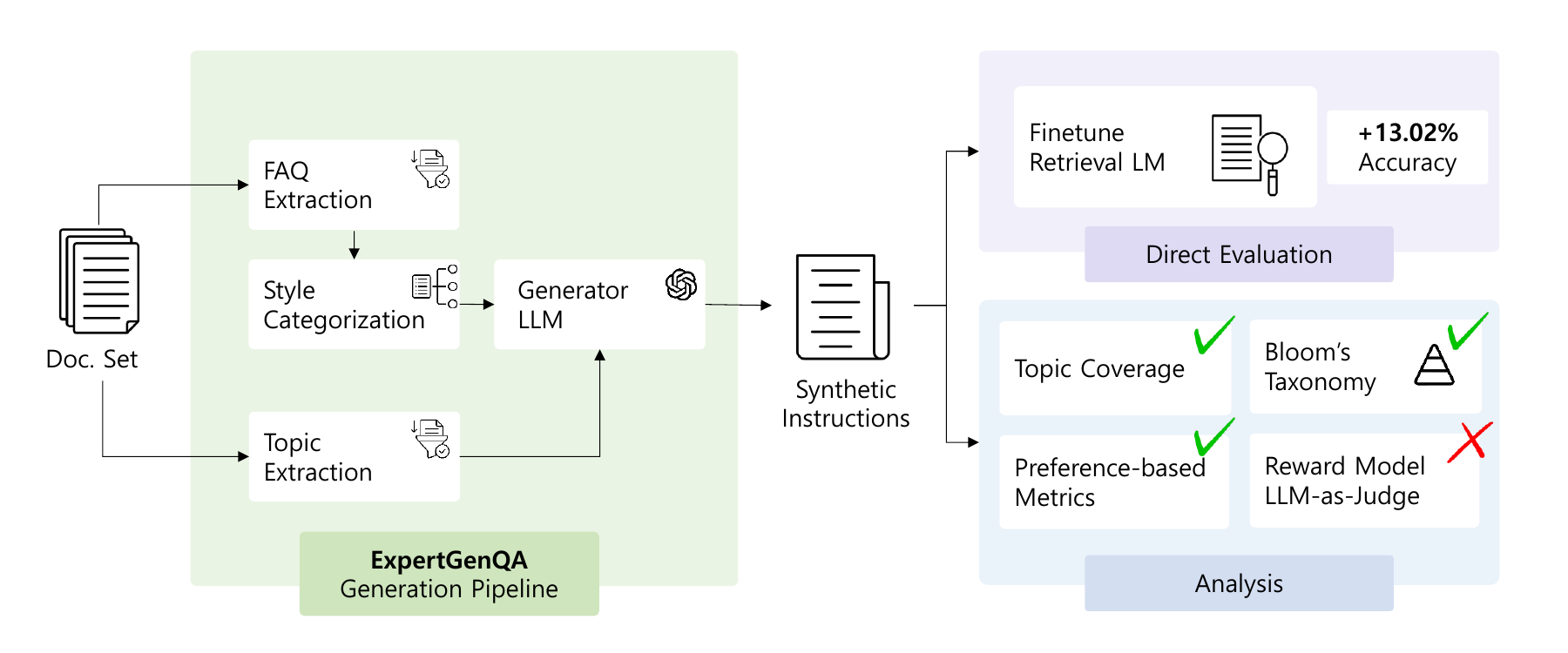}
    \caption{Overview of the ExpertGenQA pipeline (left) and proposed evaluation metrics (right). Green checkmarks (\greencheck) indicate interpretable metrics that correlate with improved retrieval accuracy, our primary evaluation metric. The red cross (\redcross) indicates our finding that both Reward Models and LLM-as-Judge show bias toward superfluous writing style and lack correlation with retrieval accuracy.}
    \label{fig:open-fig}
\end{figure*}


Generating high-quality, domain-specific questions is essential for applications such as information retrieval, reading comprehension, and knowledge assessment \citep{examiner, Kale2024FAQGenAA, e5-mistral, nvmed}. Well-crafted questions not only evaluate knowledge but also emphasize key information that domain experts consider fundamental for effective learning. They also play a critical role in document retrieval, where the quality of training queries significantly impacts model performance. More diverse, information-rich questions expose models to broader semantic patterns, enhancing their ability to generalize to unseen human-written queries \cite{e5-mistral}.

However, manually crafting such questions requires substantial domain expertise and time investment, making automatic question generation an attractive solution. While significant efforts have been made to adapt Large Language Models (LLMs) for domain-specific applications—such as BloombergGPT \citep{bloomberggpt}, FinGPT \citep{fingpt}, EcomGPT \citep{ecomgpt}, BioGPT \citep{biogpt}, and Med-PaLM \citep{medpalm1, medpalm2}—these models predominantly focus on \textit{answering} questions like an expert rather than \textit{generating} questions that imitate domain experts.

Approaches like Med-Prompt \citep{medprompt} demonstrate that domain specialization can sometimes be achieved through prompt engineering alone, simply by asking the right questions, often outperforming fine-tuned models on standard medical benchmarks. However, generating high-quality, domain-specific questions remains underexplored. Current models often default to generic, surface-level prompts \citep{Liu2024MDCureAS} that fail to incorporate the depth of expert-crafted questions. For instance, a legal professional gains little from a simplistic query like, \textit{"What is a subpoena?"} when their work demands complex, scenario-driven questions that synthesize information from statutes, precedents, and regulatory frameworks.

Moreover, assessing the domain usefulness of generated questions remains a persistent challenge. Although LLMs have advanced significantly in evaluating generated answers, existing methods often fall short when applied to question evaluation.
Widely used metrics like Reward Models (RMs) \citep{ouyang2022training, nemotron1, nemotron2, liu2024skywork} and "LLM-as-Judge" approaches \citep{llm-judge} tend to prioritize superficial features such as fluency and syntactic correctness over the semantic depth and task relevance crucial for effective document retrieval. As a result, questions that score highly according to LLM judges frequently perform poorly in domain-specific downstream retrieval tasks, failing to meet the practical needs of domain experts.

We introduce a novel question generation pipeline that learns to produce domain-specific questions by imitating a small set of expert-written examples in specialized fields. Our approach focuses on generating question-answer pairs that are not only \textbf{\textit{comprehensive in topic coverage}} but also \textbf{\textit{capture the cognitive complexity and practical needs}} of domain experts. To achieve this, we ground our method in expert-written FAQs, using them as exemplars. Our proposed pipeline, \textbf{ExpertGenQA} (Figure \ref{fig:open-fig}), employs a novel dual-categorization strategy (by style and topic). Compared to questions generated by standard few-shot prompting and template-based methods (e.g. MDCure \citep{Liu2024MDCureAS}), our domain-specific questions significantly improve document retrieval accuracy on human-written test queries, increasing top-1 accuracy from $23.96\%$ to $36.98\%$. Our detailed analysis shows that these improvements stem from diversity, cognitive load \citep{bloom, bloom_revised}, and topic coverage, all of which strongly correlate with better retrieval performance. Specifically, ExpertGenQA doubles the efficiency of baseline few-shot methods while maintaining $94.4\%$ topic coverage.

\paragraph{Why Retrieval as an Evaluation Metric?}
LLM-based evaluation metrics often reward surface-level fluency over task relevance. In contrast, retrieval performance offers a task-grounded measure of question quality, reflecting real-world utility \citep{commercialRAG}. High-quality, domain-specific questions provide better training data for the retriever, improving performance on human-written queries. We use retrieval as the primary evaluation metric, supported by auxiliary studies to analyze how diversity, cognitive load, and topic coverage contribute to these improvements.


Our contributions are:
\begin{itemize}
    \item We propose a novel question generation pipeline that leverages a small set of expert-written questions to enhance domain-specific tasks, demonstrated by its significant outperformance of baseline question generators in retrieval performance.

    \item We provide an empirical investigation into how question characteristics such as diversity, cognitive load, and topic coverage influence retrieval performance.
    \item We conduct a critical analysis revealing the bias of LLM-based judges toward superficial linguistic patterns and highlighting the need for task-relevant evaluation methods.

\end{itemize}

\section{Related Work}

\subsection{QA Generation}

Due to the growing importance of instruction tuning, the automated large-scale generation of high-quality QA has emerged as an alternative to the high cost of human curation. A straightforward approach for generating questions and answers is simply prompting pretrained LLMs \citep{wang2023selfinstructaligninglanguagemodels, stanfordalpaca, peng2023vicuna, koala_blogpost_2023}. However, LLM-generated questions often lack diversity \citep{Chen2024GenQAGM} and are prone to hallucinating facts \citep{Zhao2023HallucinationDF}.

To address these limitations, more advanced prompt-based generation approaches have been proposed. \citet{xu2024magpie} suggests a method that fully relies on the model to generate user queries without providing a seed prompt, which can otherwise restrict diversity. GenQA \citep{Chen2024GenQAGM} employs generator prompts to guide the model to first produce a broad range of topics and select one at random. MDCure \citep{Liu2024MDCureAS}, enhances complexity by prompting the model to synthesize information from multiple documents, in contrast to previous methods that primarily focused on single-document tasks. \citet{wizardlm} and \citet{mukherjee2023orca} have proposed augmentation techniques that modify questions and/or answers in existing datasets using LLMs to further scale and increase complexity.

For domain-specific requirements, there are numerous domain-specific models \citep{bloomberggpt, fingpt, ecomgpt, biogpt, medpalm1, medpalm2}, which primarily focus on answering questions rather than generating questions. Widely used domain-specific QA datasets, such as MAmmoTH \citep{mammoth} and PubMedQA \citep{jin2019pubmedqa}, are also curated manually by experts, crowdsourced, or compiled from smaller datasets.

Some studies have shown that \textit{well-crafted} QA datasets and instruction-tuned models can substantially improve performance on downstream retrieval tasks. As \citet{zhu2024inters} notes, naïve prompt-based methods or generic instruction tuning fail to capture specialized query intent, document relevance, and complex query–document relationships that are rare in general pretraining data. Interestingly, empirical evidence suggests that LLMs' instruction-following nature and information retrieval capabilities can mutually enhance each other's performance \citep{weller2024followir, wang2024instructretro}, even in domains-specific applications \citep{tran2024bioinstruct}.

\subsection{Instruction Evaluation}
Reward models (RM), which predict human preferences based on predefined criteria, are widely used as proxies for human judgment to align LLMs with human instructions \citep{ouyang2022training}. \texttt{Nemotron-70B} \citep{nemotron1, nemotron2} and \texttt{Skywork-2-27B} \citep{liu2024skywork} are notable examples of state-of-the-art RMs. Despite their effectiveness in generalized tasks, these models struggle to capture subtle language nuances and often lack training in highly specialized domains as noted by \citet{nemotron1}.

An alternative approach, LLM-as-a-Judge \citep{llm-judge, wang2024djpo}, directly utilizes LLM-generated responses for evaluation. This method enables explainable evaluations and has the potential to leverage state-of-the-art models like GPT-4o. However, these models generally underperform compared to classifier-based reward models, as they are not explicitly optimized for reward assignment \citep{Tan2024JudgeBenchAB, lambert2024rewardbenchevaluatingrewardmodels}.
\section{Methodology}
This work aims to generate synthetic questions in specialized domains that have practical utility for domain experts. To achieve this, we collect a small set of expert-written questions to serve as exemplars. Our ExpertGenQA pipeline learns domain-specific patterns from these examples and generates new questions that closely align with expert-written questions while maintaining high diversity and comprehensive coverage of source documents.

\paragraph{Data Collection}
Railway safety is critical to U.S. infrastructure, with 28\% of freight transported by rail \footnote{\url{https://www.aar.org/industries-we-support/}}. As a specialized and highly technical domain that has seen limited applications of AI, it provides an ideal test case for our approach. We select the regulatory documents published by the U.S. Federal Railroad Association (FRA), the primary federal agency that enforces safety standards and regulations on the highly decentralized and privatized U.S. rail industry.

We built our corpus by collecting 43 documents from the FRA's digital library that contain nationwide railroad regulations and guidelines (totaling 1,158 pages)\footnote{\url{https://railroads.dot.gov/elibrary-search}}. We converted the PDFs to text using the pymupdf4llm\footnote{\url{https://pypi.org/project/pymupdf4llm/}} Python package and removed pages that were primarily non-textual (containing tables, images, diagrams, etc.).

We extract 147 expert-written QAs from FAQ sections within these documents. Since the original FAQ sections did not include citations to relevant document sections, we manually identified and extracted the specific document passages that contained the information needed to respond to each question. Further details about the domain and selection criteria are provided in Appendix \ref{app:fra}.

\subsection{ExpertGenQA: A Protocol for Diverse Question Generation}


\begin{algorithm*}[!t]
\caption{ExpertGenQA Framework}
\SetKwInOut{Input}{Input}
\SetKwInOut{Output}{Output}

\Input{Document chunks $\mathcal{D}$, question styles $\mathcal{S}$, human QA pairs $\mathcal{H}$, no. few-shot combinations per style $K$, no. few-shot examples $n$}
\Output{Generated question set $\mathcal{G}$}

$\mathcal{G} \gets \emptyset$

\For{$d \in \mathcal{D}$}{
    $\mathcal{T} \gets \textsc{ExtractTopics}(d)$ \tcp*{LLM-based topic extraction (\ref{app:expertgenqa_topic_extract_prompt})}
    
    \For{$s \in \mathcal{S}$}{
        \For{$k = 1,\ldots,K$}{
            $\mathcal{F} \gets \textsc{SampleFewShot}(\mathcal{H}, s, n)$ \tcp*{Sample $n$ style-specific examples}
            \For{$t \in \mathcal{T}$}{
                $q \gets \textsc{Generate}(d, t, \mathcal{F})$ \tcp*{LLM generation with examples (\ref{app:expertgenqa_gen_prompt})}
                $\mathcal{G} \gets \mathcal{G} \cup \{(d, q)\}$
            }
        }
    }
}
\Return $\mathcal{G}$
\label{alg:instruction_generation}
\end{algorithm*}
While recent instruction generation approaches like GenQA \citep{Chen2024GenQAGM}, MDCure \citep{Liu2024MDCureAS}, and Persona Hub \citep{persona_synthetic} use template-based prompting for diversity, they sacrifice the benefits of few-shot learning and lack validation against expert instructions in technical domains. Few-shot prompting offers a straightforward method to utilize human-written QA examples \citep{brown2020language}, and we implement it as a strong comparison baseline by randomly selecting $n$ examples from our expert QA pool for each generation attempt to encourage diverse generations. However, few-shot prompting does not guarantee diverse generations or good coverage of the set of documents.

To address these limitations, we introduce a dual-categorization protocol that maintains few-shot compatibility while promoting diversity. Our method categorizes questions along two independent axes: style and topic.



\paragraph{Style Categorization} We categorize questions by style to enable focused example selection. By grouping similar questions together and using examples from just one group at a time, we can better guide the LLM to generate questions in that specific style.

We manually classify 147 expert-written questions into broad style categories. Using broad categories, rather than narrow domain-specific ones, allows the LLM to generate questions of a particular style across any arbitrary document. 

We discovered three broad categories: Policy application, which addresses how specific regulations should be interpreted; Scenario-based, which presents specific situations requiring regulatory guidance; and Terminology clarification, which focuses on defining and explaining technical terms. Analysis of our dataset revealed these categories emerged consistently across different regulatory topics, suggesting they represent fundamental question types in technical regulatory domains. Qualitative examples are provided in Appendix \ref{app:qualitative_examples}.



\paragraph{Topic Extraction} LLMs tend to focus on the most salient or interesting parts of a document when generating questions, which can result in redundant questions covering only limited portions of the text \citep{Liu2024MDCureAS}.

To address this, we use LLM-based topic extraction to identify main topics within document sections and map questions to these topics. This approach enables systematic coverage of all topics in a document.

\paragraph{Question Generation}
ExpertGenQA follows a hierarchical process to generate diverse questions. For each document $d$, it first extracts relevant topics $\mathcal{T}$ using an LLM. Then, for each question style $s$ in the style set $\mathcal{S}$, the system creates $K$ different combinations of $n$ few-shot examples. These examples are sampled from existing human QA pairs $\mathcal{H}$. For each combination of few-shot examples and each extracted topic, the system generates a new question using an LLM.

The pipeline's efficiency comes from its structure: by processing topics in the innermost loop, the LLM only needs to handle the few-shot examples once per combination, utilizing prefix-caching for subsequent topic-based generations.

The complete generation process is detailed in Algorithm \ref{alg:instruction_generation}. After generation, we remove near-duplicate and paraphrased questions using a bigram overlap algorithm following \citet{deduplication}.


\subsection{Retrieval Evaluation}
Standard retrieval models perform poorly in specialized technical domains like railway regulations, where documents have similar vocabulary, structure, and overlapping terminology \citep{xu2024simrag, lewis2021retrievalaugmented}. This underperformance stems from several domain-specific challenges. Technical domains use specialized vocabulary where subtle distinctions carry significant regulatory implications, and general-purpose retrievers struggle to disambiguate these nuanced differences. Additionally, regulatory documents often follow standardized formats with similar section structures and phrasing patterns, making it difficult for retrievers to distinguish between relevant and irrelevant passages. 

We leverage these inherent challenges as a robust evaluation framework. The reasoning is straightforward: synthetic questions that better capture domain expertise should lead to measurably improved retrieval performance when used as training data (with strict deduplication against the test data). This approach directly measures the downstream practical utility of synthetic data for information retrieval.

For each synthetic generation pipeline, we finetune a retrieval LM, \texttt{gte-modernbert-base} \citep{gte, modernbert} using the generated document-query pairs and evaluate performance using the human-authored QA pairs as a test set. This provides a highly practical signal for comparing different question-generation approaches based on their utility for downstream retrieval tasks. We use the InfoNCE loss \citep{infonce} to fine-tune retrieval LMs \ref{alg:infonce}. InfoNCE loss compares a positive pair of samples (like a query and its corresponding document) against multiple negative pairs, encouraging the model to maximize agreement between positive pairs while pushing apart negative pairs in the representation space.


\begin{equation}
\label{alg:infonce}
L_{infoNCE} = -\log \frac{e^{s(q,d^+)/\tau}}{e^{s(q,d^+)/\tau} + \sum_{i=1}^n c^{s(q,d_i^-)/\tau}} \tag{3}
\end{equation}

where $s(q,d)$ is the similarity function between query embedding $q$ and a document embedding $d$, $d^+$ is a document embedding relevant to answering $q$ and $\mathcal{D} = \{d_1^-, \ldots, d_n^-\}$ is a set of irrelevant document embeddings. $\tau$ is a temperature hyperparameter that controls the sharpness of the probability distribution over similarities. We use only in-batch negatives instead of mining hard negatives for simplicity \citep{nvmed}.


\section{Experimental Setup}
We employ \texttt{GPT-4o} \citep{achiam2023gpt} as our primary language model for all tasks including topic extraction, question evaluation, Bloom's Taxonomy classification, and response generation. Consistent with \citet{Chen2024GenQAGM}, all generation pipelines use temperature $T=1$ and sample 5 generations per input. We opt for a strict bigram-overlap threshold of $0.3$ for near-duplicate removal.

For retrieval evaluation, we use the state-of-the-art \texttt{NVEmbed-70B-V2} \citep{nvmed} as the zero-shot baseline and finetune \texttt{Alibaba-NLP/gte-modernbert-base} \citep{gte}, a smaller ($\sim150$M) yet capable retrieval model using batch size $64$, learning rate $1e-5$ and InfoNCE loss \citep{infonce} with temperature $T=0.1$ with only in-batch negatives. We use cosine similarity as the similarity function $s(q,d)$. We use the 147 expert-written QA as the test set and the generated questions as the train sets.

\section{Main Results}

\subsection{Diversity of Generated Questions and Pipeline Efficiency}
We evaluate the efficiency of ExpertGenQA against two baselines: few-shot prompting and MDCure \citep{Liu2024MDCureAS}, a prompt-template-based pipeline that does not use examples. Generating diverse synthetic questions is important not only for downstream applications but also for efficiency reasons, as redundant generations result in wasted LLM calls.

\begin{figure}[!h]
    \centering
    \includegraphics[width=1\linewidth]{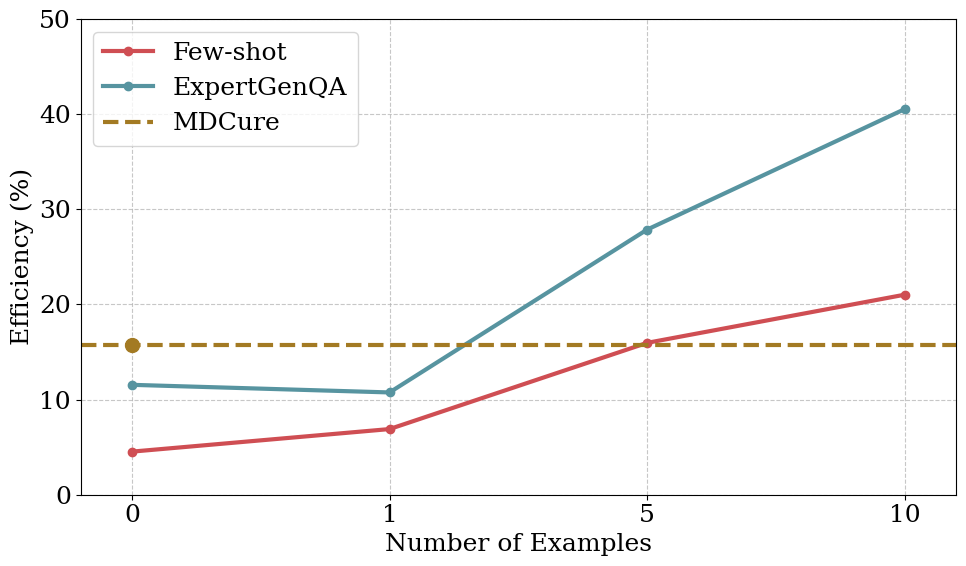}
    \caption{Comparison of efficiency across question-generation pipelines over the different number of few-shot examples. We define efficiency as the fraction of unique generations over the total sampled generations. }
    \label{fig:efficiency}
\end{figure}

Figure \ref{fig:efficiency} demonstrates that ExpertGenQA with 10 examples produces twice as many unique questions as the few-shot prompting baseline for the same number of LLM calls. 
More examples generally increase the efficiency of both ExpertGenQA and few-shot prompting. With 10 examples, ExpertGenQA generates $7,140$ unique questions from $17,622$ LLM calls, while 10-shot prompting generates only $3,658$ unique questions. In contrast, MDCure, being a purely template-based approach without examples, maintains a static efficiency of 15.71\%, generating $8,030$ instructions from $51,100$ LLM calls. The detailed statistics (e.g., the total number of samples and unique questions) of the generated questions can be found in Appendix \ref{app:diversity_efficiency}. Qualitative examples of synthetic intructions are included in Appendix \ref{app:qualitative_examples} .


\subsection{Retrieval LM}
Table \ref{tab:retreival} shows that finetuning a retrieval LM \texttt{AlibabaNLP/gte-modernbert-base} \citep{gte, modernbert} on ExpertGenQA generations significantly improves top-1 retrieval accuracy from $23.96\%$ to $36.98\%$, outperforming even the much larger generalist retrieval LM NVEmbed-V2 \citep{nvmed}. In contrast, finetuning on synthetic instructions from 10-shot prompting and MDCure yields more modest improvements of $+7.15\%$ and $+3.64\%$ respectively. Notably, the retrieval LM fine-tuned on MDCure-generated data achieves lower retrieval accuracy than the 10-shot pipeline, despite having more than twice the training data (8,030 instructions vs. 3,658). This demonstrates the importance of synthetic data matching the complexity and utility of expert-written QA for practical applications like retrieval.

\begin{table}[!h]
    \centering
    \begin{tabular}{llcc}
       \toprule
       Model  &  Param&Top-1 & Top-5 \\
       \midrule
       \texttt{gte-baseline} &  150M & 23.96 & 55.73\\
       \texttt{NVEmbed-V2}& 7B & 29.17 & 60.94\\
       \midrule
       \texttt{gte[MDCure]}&  150M & 27.60 & 53.65\\
       \texttt{gte[10-Shot]}&  150M & 31.11 & 60.50\\
       \texttt{gte[ExpertGenQA]} &  150M & \textbf{36.98} & \textbf{77.08}\\
       \bottomrule
    \end{tabular}
    \caption{Retrieval performance (in terms of Top$-k$ accuracy) of retrieval LMs and finetuned variants. The second column contains the number of parameters. \texttt{gte[X]} means the \texttt{AlibabaNLP/gte-modernbert-base} was fine-tuned on synthetic instructions from the respective dataset generated using the \texttt{X} pipeline. Both \texttt{few-shot} and \texttt{ExpertGenQA} pipelines use 10 examples. }
    \label{tab:retreival}
\end{table}


\section{Analysis}

While directly finetuning a retrieval language model provides the most accurate measure of synthetic question effectiveness, it is impractical for regular use during pipeline development due to the compute required. Therefore, we investigate alternative evaluation metrics to determine which ones meaningfully correlate with improvements in retrieval language model performance. These metrics could serve as more practical proxies for assessing question quality during the development process.

\subsection{Reward Models and LLM-as-Judge}

\begin{figure}[!h]
\centering
\includegraphics[width=1\linewidth]{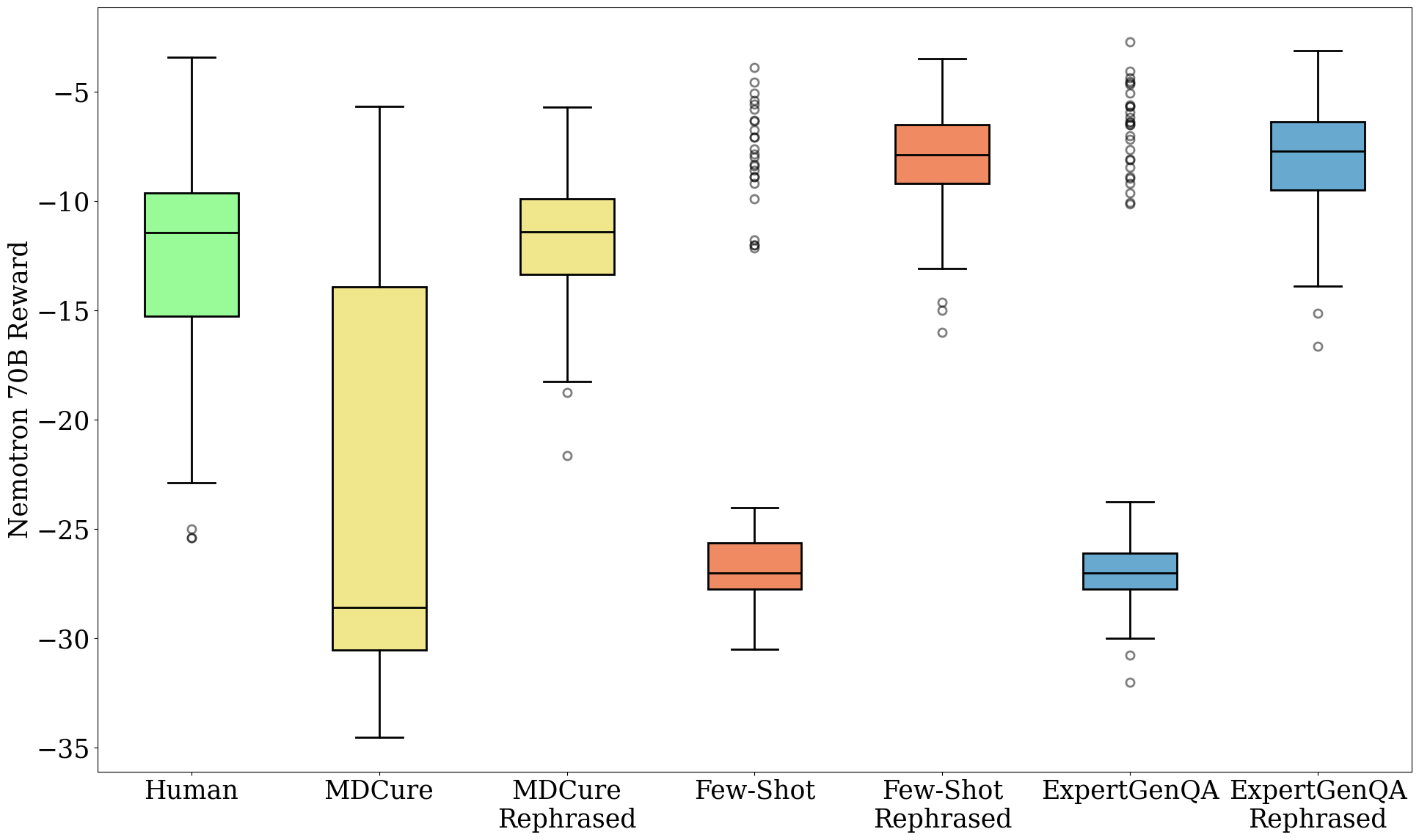}
\caption{Box plot of reward assigned to  questions by \texttt{Llama-3.1-Nemotron-70B-Instruct} Reward Model.  Notably, merely rephrasing synthetic questions to sound \textit{human-like} drastically increases the assigned reward score although the semantic content hasn't changed.}
\label{fig:reward_histogram}
\end{figure}


We test the ability of state-of-the-art Reward Models (RM) to judge question quality, based on which we compare the quality  of expert-written questions with synthetically generated questions using the template shown in Appendix \ref{app:reward_prompt}. We leveraged GPT4o to automatically rephrase synthetic questions that sound human-like using the prompt in Appendix \ref{app:paraphrase_prompt}. In Fig. \ref{fig:reward_histogram}, we demonstrate the reward scores of human-written questions (see "Human"), synthetic questions using the aforementioned three pipelines, and synthetic questions after rephrasing ( see "X Rephrased" where "X" is the corresponding question generation pipeline). Clearly, 1) merely rephrasing LLM generations drastically increases the score awarded by the RM; and 2) synthetic generations with rephrasing achieve higher rewards than expert-written questions.  Thus, the results in Fig. \ref{fig:reward_histogram} imply that \texttt{Nemotron-70B-Instruct} RM \citep{nemotron1, nemotron2}  exhibits a strong bias based on writing style rather than content quality.  We also observe such  bias in another state-of-the-art RM \texttt{Skywork-Reward-Gemma-2-27B-v0.2} \citep{liu2024skywork} while using \textit{GPT4o-as-Judge} (see the details results in Appendix \ref{app:judge_reward}). These findings indicate that RMs are highly sensitive to superficial stylistic changes and \textbf{do not correlate} with the clear differences between different pipelines shown in Table \ref{tab:retreival}.

\subsection{Cognitive Complexity Distribution} 

To evaluate the cognitive complexity and educational value of generated questions, we leverage Bloom's Revised Taxonomy \cite{bloom, bloom_revised}, a well-established framework from cognitive science that categorizes learning objectives into six hierarchical levels: Remember, Understand, Apply, Analyze, Evaluate, and Create. Each level represents increasingly complex cognitive processes, from basic recall to sophisticated synthesis. We use GPT4o to classify both human-written and synthetic questions according to these taxonomic levels, allowing us to assess and compare the distribution of cognitive demands across different instruction sets.

\begin{figure}[!h]
    \centering
    \includegraphics[width=1\linewidth]{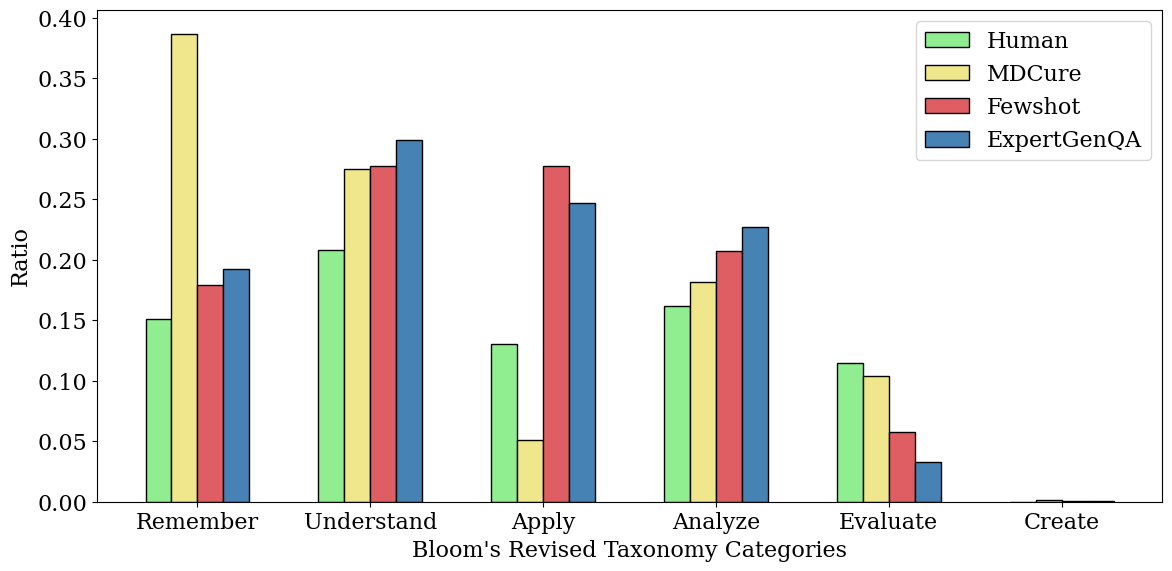}
    \caption{Distribution of cognitive complexity levels in human-written and synthetic instructions according to Bloom's Revised Taxonomy. MDCure shows higher concentration in lower cognitive levels.}
    \label{fig:bloom_barplot}
\end{figure}

Fig. \ref{fig:bloom_barplot} shows the distribution of instructions across Bloom's Taxonomy levels for human-written and synthetic data. MDCure shows a notable skew toward lower-level cognitive tasks, with approximately $39\%$ of instructions falling into the \textit{Remember} category. The distribution of human-written questions demonstrates greater uniformity across cognitive levels, reflecting their origin from domain experts crafting questions for other domain experts. Few-shot prompting and ExpertGenQA produce distributions more closely aligned with human-written questions, emphasizing the value of incorporating few-shot examples in specialized domains. 


\subsection{Topic Coverage and Preference Metrics}

A key challenge in question generation is ensuring comprehensive coverage of source materials, as missing critical topics could lead to gaps in downstream capabilities. To ensure that generated questions span the full scope of document content, we measure topic coverage:

\begin{equation}
TC = \frac{1}{|\mathcal{D}|} \sum_{d \in \mathcal{D}} \frac{|Q(d)|}{|T(d)|}
\end{equation}

\noindent where $Q(d)$ represents the topics covered by generated questions for document $d$, $T(d)$ represents the topics in document $d$, $\mathcal{D}$ is the document set, and $|\cdot|$ is the set cardinality operator.

Reward models are generally trained to evaluate responses to questions rather than the questions themselves. \citet{yu2025rip} has shown that the rewards assigned to responses can be used as a proxy metric for instruction quality. Following their methodology, we sample $N=10$ responses for each context-question $x$ and evaluate them using a RM. From these responses, we identify the \textit{chosen} response $y_w$ with the highest reward and the \textit{rejected} response $y_l$ with the lowest reward. We analyze three key metrics:

\begin{itemize}
\setlength\itemsep{0em}
    \item Rejected response reward: $\texttt{RM}(y_l|x)$; higher is better
    \item Rejected response length ratio: $\frac{\texttt{len}(y_l)}{\texttt{len}(x)}$; higher is better
    \item Reward gap: $\Delta\texttt{RM}(\cdot) = \texttt{RM}(y_w|x) - \texttt{RM}(y_l|x)$; lower is better
\end{itemize}

The intuition behind these metrics, as demonstrated by \citet{yu2025rip}, is that high-quality instructions should produce longer and more coherent responses even when they are ``rejected" and should lead to more consistent response quality.

\begin{table}[h]
\centering
\resizebox{\columnwidth}{!}{
\begin{tabular}{lccccc}
\toprule
Model & $\#$Topics & {$TC\uparrow$} & {$\texttt{RM}(y_l|x)\uparrow$} & {$\frac{\texttt{len}(y_l)}{\texttt{len}(x)}\uparrow$} & {$\Delta\texttt{RM}(\cdot)\downarrow$} \\
\midrule
MDCure & $8,030$ & 0.626 & -8.67 & 0.27& \textbf{4.38} \\
Few-Shot & $3,658$ & 0.726 & -7.87 & 0.59& 5.15 \\
EGenQA & $7,140$ & \textbf{0.944} & \textbf{-7.75} & \textbf{0.61}& 5.05 \\
\bottomrule
\end{tabular}
}
\caption{Comparison of pipelines across topic coverage ($TC$) and response preference metrics. ExpertGenQA (EGenQA) achieves the best scores in $TC$, rejected response quality $\texttt{RM}(y_l|x)\uparrow$, and rejected response length ratio  $\frac{\texttt{len}(y_l)}{\texttt{len}(x)}\uparrow$.}.
\label{tab:new_metrics}
\end{table}

The quantitative evaluation in Table \ref{tab:new_metrics} reveals interesting trade-offs between the three approaches. ExpertGenQA has the highest rejected response reward, rejected response length ratio, and topic coverage even after filtering, highlighting the effectiveness of the ExpertGenQA generation protocol. Further investigation into the reward gap $\Delta\texttt{RM}(\cdot)$ reveals an interesting pattern when analyzed alongside Bloom's Taxonomy levels, as shown in appendix \ref{app:response_based_rewards} table \ref{tab:bloom_v_reward}. While MDCure achieves the lowest reward gap despite generating simpler questions, this appears to be a natural consequence of its approach - simpler questions tend to elicit more consistent responses from LLMs, resulting in smaller reward gaps. From the four metrics, topic coverage strongly correlates with the retrieval performance in Table \ref{tab:retreival}.
\section{Conclusion}
This work demonstrates that combining structured categorization with few-shot learning can effectively generate domain-specific questions for railway regulations. Our evaluation reveals limitations in current automated assessment methods, as both reward models and LLM-as-judge approaches struggle to meaningfully evaluate technical content quality. The cognitive complexity analysis shows our approach better preserves the distribution of expert-level thinking demands compared to template-based methods.


Importantly, our generated questions achieved improved retrieval performance (+13.02\% top-1 accuracy) compared to the best competing model, although the modest absolute performance (36.98\%) highlights ongoing challenges in technical domain retrieval. In the future, we will extend this approach to other specialized fileds where expert knowledge is crucial but limited.



\section{Limitations}
Our study has several limitations. Firstly, we focused exclusively on the Federal Railway Administration domain because it offered a well-structured corpus of regulatory documents with expert-written FAQs, making it an ideal testing ground for our approach. However, Our proposed pipeline ExpertGenQA and evaluation metrics should be effective in other specialized domains as well. We leave this as future work. Secondly, while ExpertGenQA significantly improves retrieval performance compared to baselines, the best top-1 accuracy remains below 40\%. Scaling up synthetic data generation is a promising direction for achieving practically viable performance levels.
Finally, the few-shot prompting component of ExpertGenQA, while effective for quality, incurs substantial compute costs in terms of token usage during generation. Future research could explore optimizing the efficiency-quality tradeoff.
\section{Ethical Considerations}

This work focuses on generating high-quality question-answer pairs for specialized technical domains.  We acknowledge the following ethical considerations:

\begin{itemize}
    \item \textbf{Data Source and Copyright:}  We used publicly available U.S. Federal Railroad Administration (FRA) documents as a case study. While these documents are in the public domain, it's important to recognize that not all information on the internet is free for unrestricted use. In this work, we processed the PDF documents using the \texttt{pymupdf4llm} library, adhering to its intended use and licensing terms. We encourage future research to carefully consider data provenance and usage rights when extending this methodology to other domains.

    \item \textbf{Risk of Data Poisoning:}  While our current work uses a curated set of official FRA documents, extending this approach to less controlled environments introduces the risk of data poisoning.  Malicious actors could intentionally introduce incorrect or misleading information into the source documents used for question generation.  This could lead to the generation of inaccurate or biased question-answer pairs, ultimately impacting the reliability of downstream applications like retrieval systems. 

    \item \textbf{Ensuring Trustworthy Information:}  The primary goal of this work is to improve information access and knowledge assessment for domain experts.  However, there is a risk that errors in the generated questions or retrieved information could lead to incorrect conclusions or decisions by these experts.  Ensuring the accuracy and reliability of the generated content is crucial for building trustworthy AI systems. 
\end{itemize}

We believe that the benefits of this research, particularly in providing more efficient access to critical information in specialized domains, are substantial.  However, we emphasize the importance of responsible development and deployment, with careful consideration of data quality, potential risks, and the need for ongoing validation to ensure trustworthy and reliable results.

\section{Acknowledgements}
This work was supported by the University Transportation Center for Railway Safety (UTCRS) at UTRGV through the USDOT UTC Program under Grant No. $69A3552348340$ and NSF CREST Center for Multidisciplinary Research Excellence in Cyber-Physical Infrastructure Systems (MECIS) grant no. $2112650$. Any opinions, findings, conclusions, or recommendations expressed in this material are those of the author(s) and do not necessarily reflect the views of the funding parties.

\bibliography{custom}

\appendix
\onecolumn

\section{Federal Railway Administration}
\label{app:fra}
The U.S. railway system operates primarily under private ownership, with freight railroads owned and operated by corporations such as Union Pacific, BNSF, and CSX. These companies handle a significant portion of the nation's freight transportation, moving over a third of goods by ton-miles. Passenger rail services, which are much more limited in scope, include Amtrak (a federally supported corporation) and various regional commuter systems such as Metrolink, BART, and SEPTA. Most passenger rail services operate on infrastructure owned and maintained by private freight railroads, creating a complex system of shared use that requires extensive oversight and coordination.

The Federal Railroad Administration (FRA), part of the U.S. Department of Transportation (USDOT), serves as the primary federal agency responsible for regulating and supporting this privately managed rail system. The FRA's role includes developing and enforcing safety standards for infrastructure, rail equipment, operations, and employee working conditions. Its inspectors ensure compliance across the industry, enforce safety mandates, and investigate accidents to improve future practices.

In addition to enforcing safety standards, the FRA administers funding programs, such as the Consolidated Rail Infrastructure and Safety Improvements (CRISI) grant initiative, which supports infrastructure modernization, capacity improvements, and the implementation of new technologies. The agency also collects and distributes data on accident trends, track performance, and operator compliance, providing essential insights for railroads, policymakers, and the public to guide decision-making and planning.  The FRA maintains an extensive online repository of regulatory and informational documents through its eLibrary to support industry stakeholders, researchers, and the public. The eLibrary \footnote{\url{https://railroads.dot.gov/elibrary-search}} contains more than 9000 documents spanning from 1966 to the present.

In this work, We have curated 43 up-to-date documents from the FRA eLibrary based on the following criteria: sufficient textual content and expert-written QA pairs that are not tied to specific events or overly focused on temporary or local programs. A significant portion of the qualified QA pairs comes from the \textit{Federal Railroad Administration Guide for Preparing Accident/Incident Reports}, which provides comprehensive regulatory explanations and practical QA pairs for each section. Additional sources include FAQs and QA-focused documents covering topics such as workers, programs, operations, and services. Examples include \textit{Questions and Answers Concerning Wheelchairs and Bus and Rail Service} and \textit{RCL Operations Q\&As}.

\section{Diversity and Efficiency}
\label{app:diversity_efficiency}

\begin{table}[h]
    \centering
    \begin{tabular}{l|cccc}
        \toprule
        Strategy & \#Shots & \#LLM() & \#Unique & Efficiency\\
        \midrule
        MDCure & 0 & 51,100 & 8,030 & 15.71\% \\
        \midrule
        \multirow{4}{*}{Few-shot} & 0 & 17,400 & 788 & 4.53\% \\
        & 1 & 17,400 & 1,220 &  6.90\%\\
        & 5 & 17,400& 2,778& 15.96\%\\
        & 10 & 17,400 & 3,658 & 21.02\% \\
        \midrule
        \multirow{4}{*}{ExpertGenQA} & 0 & 17,622 & 2,035 & 11.55\%  \\
         & 1 & 24,030 & 2,584 & 10.75\%\\
         & 5 & 19,224 & 5,355 & 27.86\%\\
         & 10 & 17,622  & 7,140 & 40.52\% \\
        \bottomrule
    \end{tabular}
    \caption{Efficiency of different generation pipelines. \#Shots denotes the number of few-shot examples used, \#LLM() is the number of LLM calls, \#Unique is the number of questions left after deduplication, and Efficiency denotes the ratio of unique questions over total LLM calls used.}
    \label{tab:efficiency}
\end{table}

\section{Data Generation with MDCure}

MDCure \citep{Liu2024MDCureAS} is a pipeline for generating question-answers from single or multiple documents in a zero-shot setting. After generation, it uses the MDCure Reward Model 
(MDCureRM) to filter the generations. MDCure uses three categories of prompts to encourage generation diversity: generic, template-based, and snippet-based. MDCure first clusters documents by their embeddings. For each cluster, generic prompts ask the model to generate questions requiring all the cluster documents to answer. Template-based prompts are constructed by randomly combining restrictions on the question and answer such as question type (summarization, paraphrasing, inference, etc.), answer length, and question style (declarative, imperative, etc.). Finally, snippet-based prompts work on similar pairs of documents instead of clusters. MDCure first extracts random snippets from each document and prompts a model to generate a question and answer based on the two snippets.

We generate 5170 QAs using generic prompts, 14300 with template-based prompts, and 31080 with snippet-based prompts from our FRA documents. For a fair comparison with ExpertGenQA, We sample 5 completions per prompt. We use MDCureRM to score the generations and keep the top 50\% generations by score. Similar to ExpertGenQA, we further filter near-duplicates by word overlap. The complete pipeline yields 8030 QA pairs from 51100 sampled generations.

\section{Reward Models and LLM-as-Judge}
\label{app:judge_reward}

\begin{table*}[!h]
\centering
\begin{tabular}{l|cccc}
\toprule
& Human & MDCure & FewShot & ExpertGenQA \\
\midrule
Relevance & 4.44 & 4.18 & 4.49 & \textbf{4.48} \\
Coherence/Factuality & 4.23 & 4.19 & \textbf{4.33} & 4.31 \\
Creativity & 2.99 & 3.13 & 3.11 & \textbf{3.16} \\
Context Integration & 3.32 & 3.47 & \textbf{3.48} & 3.43 \\
Intra-doc Relations & 3.62 & 3.58 & \textbf{3.81} & 3.66 \\
Complexity & 3.32 & 3.46 & 3.44 & \textbf{3.52} \\
\bottomrule
\end{tabular}
\caption{Fine-grained scores assigned by  \texttt{GPT4o-as-Judge} using the MDCure prompt \citep{Liu2024MDCureAS}. The best score for each metric is in bold. The weighted-average score is shown in Fig. \ref{fig:gpt4o-boxplot}. We use the weights proposed by MDCure.}
\label{tab:llm-as-judge-scores}
\end{table*}

\begin{figure*}[!h]
    \centering
    \includegraphics[width=0.7\linewidth]{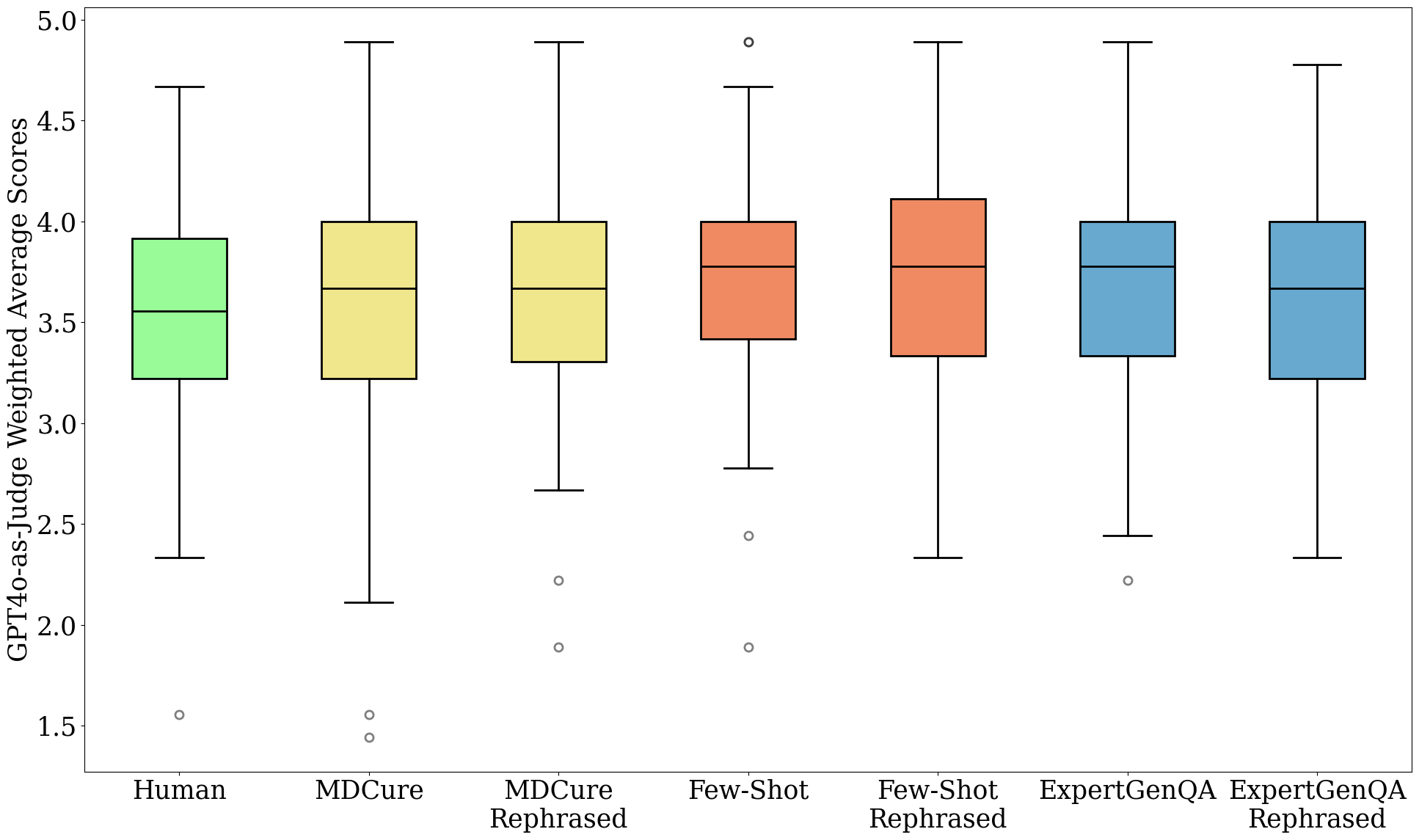}
    \caption{Box plot of scores assigned by \texttt{GPT4o-as-Judge} using the MDCure prompt \citep{Liu2024MDCureAS}.  \texttt{GPT4o-as-Judge} assigned similar scores for all generation methods and hence does not correlate with the clear differences in downstream task improvements shown in Table \ref{tab:retreival}.}
    \label{fig:gpt4o-boxplot}
\end{figure*}

\begin{figure*}[!h]
    \centering
    \includegraphics[width=0.7\linewidth]{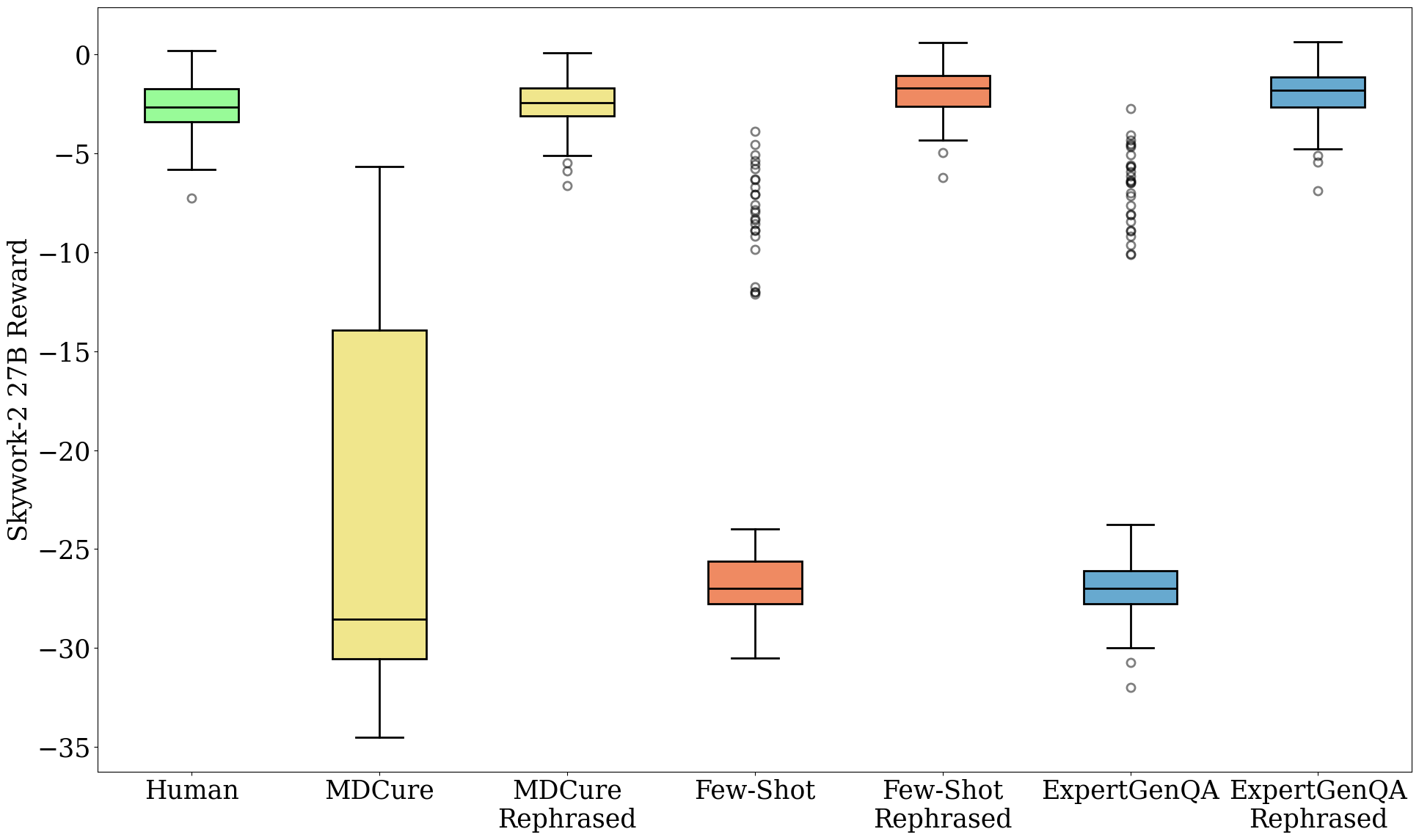}
    \caption{Box plot of reward assigned by \texttt{Skywork-Reward-27B} Reward Model. Merely rephrasing synthetic instructions to sound \textit{human-like} drastically increases the assigned reward showing that RMs are not suitable for judging synthetic instruction quality.}
    \label{fig:skywork-boxplot}
\end{figure*}

\section{Evaluation via Response Generation}
\label{app:response_based_rewards}

\begin{table*}[htbp]
\centering
\begin{tabular}{l|ccc|ccc|ccc}
\toprule
\multirow{2}{*}{\textbf{Level}} & \multicolumn{3}{c|}{$\mathbf{RM(y_l|x)}\uparrow$} & \multicolumn{3}{c|}{$\frac{\mathbf{len(y_l)}}{\mathbf{len(x)}}\uparrow$} & \multicolumn{3}{c}{$\mathbf{\Delta RM(\cdot)\downarrow}$} \\
& MD & FS & EX & MD & FS & EX & MD & FS & EX \\
\midrule
Remember & \textbf{-7.77} & -8.29 & -7.84 & 0.20& \textbf{0.56}& 0.40& \textbf{3.79}& 5.23 & 4.62\\
\midrule
Understand & -9.66 & \textbf{-6.89} & -7.52 & 0.31& 0.55& \textbf{0.59}& 5.00 & \textbf{4.03} & 4.84 \\
\midrule
Apply & -9.52 & -8.21 & \textbf{-7.77} & 0.35& 0.66& \textbf{0.80}& 6.08 & 5.56 & \textbf{5.40} \\
\midrule
Analyze & \textbf{-7.37} & -8.26 & -8.23 & 0.35& 0.59& \textbf{0.69}& \textbf{3.23} & 5.83 & 5.76 \\
\midrule
Evaluate & -11.55 & -8.50 & \textbf{-5.91} & 0.17& \textbf{0.56}& 0.44& 6.03 & 6.23 & \textbf{3.16} \\
\midrule
\midrule
 Average & -8.67 & -7.87 & \textbf{-7.75} & 0.27 & 0.59 & \textbf{0.61}& \textbf{4.38} & 5.15 & 5.05\\
\bottomrule
\end{tabular}

\caption{Comparison of response preference metrics against Bloom's Taxonomy. MD: MDCure, FS: FewShot, EX: ExpertGenQA. The best performance is in bold.}
\label{tab:bloom_v_reward}
\end{table*}

\vspace{3in}

\section{Qualititative Examples}
\label{app:qualitative_examples}

\paragraph{Expert-written Questions (Randomly Sampled)}
\begin{mdframed}
\textbf{Policy Application}

\noindent
1. Our employees are frequently tested for drug or alcohol use after an accident/incident. Company policy prohibits an employee from returning to work until the results of the tests are known and it is established that there is no risk factor due to impairment. Must we make a report because of the days the employee was held out of service while awaiting test results?

\noindent
2. Our employees are frequently tested for drug or alcohol use after an accident/incident. Company policy prohibits an employee from returning to work until the results of the tests are known and it is established that there is no risk factor due to impairment. Must we make a report because of the days the employee was held out of service while awaiting test results?

\noindent
3. How do I decide if a case is work-related when the employee is working at
home or telecommuting from another location?

\noindent
\textbf{Scenario-based}

\noindent
4. If the injured or ill worker produces fewer goods or services than he or she would have produced prior to the injury or illness, but otherwise performs all of the routine functions of his or her work, is the case considered a restricted work case?

\noindent
5. Say that a highway user struck a signal stand at a highway-rail grade crossing and was injured, but there was no on-track equipment present, nor were employees of the railroad in the vicinity. Is this reportable?

\noindent
6. One of our employees experienced minor musculoskeletal discomfort. The health care professional who examined the employee only provided first aid treatment. In addition, it was determined that the employee is fully able to perform all of her routine job functions. When the employee returned to work, we decided to limit the duties of the employee for the purpose of preventing a more serious condition from developing. Is this a restricted work case?

\noindent
\textbf{Terminology Clarification}

\noindent
7. Is a physical therapist considered a "health care professional" under the definition of health care professional?

\noindent
8. Removing splinters or foreign material from areas other than the eye by irrigation, tweezers, cotton swabs, or other simple means... What are "other simple means" of removing splinters that are considered first aid?

\noindent
9. What does "other potentially infectious material" mean?

\end{mdframed}

\paragraph{Synthetic Instructions from MDCure (Randomly Sampled)}
\begin{mdframed}
\noindent
1. How do the reporting requirements for railroad accidents and incidents ensure timely and accurate accountability while also protecting the rights of employees involved, particularly in cases where human factors are cited as a cause?

\noindent
2. What are the requirements for a written request to treat subsidiary railroads as a single, integrated railroad system, and what does railroad transportation encompass according to the regulations?

\noindent
3. What are the specific reporting criteria and procedures for railroads regarding suicide data, as well as exceptions related to injuries or illnesses incurred by employees, contractors, and volunteers?

\noindent
4. Can a person who is not on railroad property be involved in railroad operations?

\noindent
5. What are the reporting criteria for workplace injuries and the investigation procedures for rail accidents regarding substance use?

\noindent
6. Railroad injury and illness reporting conditions?

\noindent
7. What must be submitted for FRA review?

\noindent
8. How does the categorization of accidents and the reporting thresholds relate to the documentation requirements for rail equipment incidents and worker injuries within the railroad industry?

\noindent
9. Where to download FRA forms and guide?

\noindent
10. What are the primary purposes of Part 225 regulations compared to the applicability restrictions outlined in § 225.3?

\end{mdframed}

\paragraph{Synthetic Instructions from Fewshot Prompting (Randomly Sampled)}
\begin{mdframed}
\noindent
1. If an accident involves hazardous materials but no evacuation was necessary, should the number of people evacuated still be reported as "0," or is it considered not applicable?

\noindent
2. If a volunteer railroad worker is injured while performing safety-sensitive functions, does that injury require reporting under FRA regulations?

\noindent
3. If a railroad operates another company's freight train and runs a total of 1,000 miles with its crew during the month, should those miles be reported in the total for the operating railroad or the railroad that owns the freight train?

\noindent
4. In the situation where an employee broke their arm during a physical altercation with a coworker in the company parking lot before clocking in for work, is there a justification for classifying this injury as non-work-related, or must it be reported as a work-related incident?

\noindent
5. Are incidents involving damage to idle railroad cars due to vandalism by nonrailroad employees subject to reporting if there is no involvement of railroad employees?

\noindent
6. If a railroad employee suffers a reportable injury and the railroad receives information about it six days later, what is the latest date by which the railroad must enter that reportable case on the appropriate record?

\noindent
7. What should a railroad do if they receive an Employee Statement Supplementing Railroad Accident Report after initially filing the Rail Equipment Accident/Incident Report?

\noindent
8. If a railroad experiences a significant change in their reported damage costs for a rail equipment accident after initially filing a report, what is the percentage variance that would necessitate an amended report?

\noindent
9. An employee was injured when a heavy object fell on them while they were chatting with a co-worker in the break room. How should we determine if this injury is considered work-related under the FRA guidelines?

\noindent
10. Are railroads required to include suicide data in their periodic reports to FRA, and if not, how is such data handled?

\end{mdframed}

\paragraph{Synthetic Instructions from ExpertGenQA (Randomly Sampled)}

\begin{mdframed}

\noindent
1. What is the significance of the FRA Guide for Preparing Accident/Incident Reports in relation to Part 225, and how does it serve railroad companies in meeting their recordkeeping and reporting obligations?

\noindent
2. If an employee tested positive for drug use following an accident and further investigation indicates that drug use did not impair their ability to perform their job responsibilities, how should this be documented in the accident report narrative? What specific information should be included to clearly explain this determination?

\noindent
3. In the context of reporting an incident involving a highway user and railroad on-track equipment, how should a railroad handle a situation where a highway user attempted to avoid the incident but was struck at a different location than the crossing?

\noindent
4. What guidelines must be followed when determining whether a case falls under the exceptions for reporting injuries or illnesses?

\noindent
5. What types of professionals are classified as "qualified health care professionals," and what does this classification entail regarding their scope of practice?

\noindent
6. In Scenario 4, if the employee injured during a smoke break was on a designated break time and the employer has a policy allowing such breaks, would this change the work-relatedness assessment for the slip on ice, leading it to be reportable?

\noindent
7. What criteria define a "significant injury" or "significant illness" in the context of reporting railroad accidents or incidents?

\noindent
8. What information is required to be maintained in a railroad's injury and illness record, and can alternative recordkeeping formats be used?

\noindent
9. What defines occupational tuberculosis in the context of railroad employees?

\noindent
10. What are the three primary groups into which reportable railroad accidents and incidents are categorized, and what are the specific reporting requirements for each group?
\end{mdframed}

\subsection{Prompt Template}
\label{app:prompts}

\subsubsection{ExpertGenQA Topic Extraction Prompt}
\label{app:expertgenqa_topic_extract_prompt}

\begin{mdframed}
\begin{verbatim}
Passage: {{PASSAGE}}

-----

Please analyze the given passage and identify its main topics. Provide your response 
in JSON format where the key is 'topics' and its value is an array of the main 
topic names. For example:

{
'topics': ['topic1', 'topic2', 'topic3']
}
\end{verbatim}
\end{mdframed}

\subsubsection{ExpertGenQA Generation Prompt}
\label{app:expertgenqa_gen_prompt}

\begin{mdframed}
\begin{verbatim}
Passage: {{PASSAGE}}

-----

The passage above covers the following topics:
{{TOPICS_IN_PASSAGE}}

Generate a question from the passage related to '{{SELECTED_TOPIC}}'.  
\end{verbatim}
\end{mdframed}

\subsubsection{Paraphrasing with Examples - User Instruction}
\label{app:paraphrase_prompt}

\begin{mdframed}
\begin{verbatim}
<target_question>
{{QUESTION}}
</target_question>

<examples>
{{EXAMPLES}}
</examples>

Please paraphrase the target question to match the style of the examples. Do not make 
any changes that would alter the meaning and change its answer. Do not answer the 
question. Respond with only the rephrased question (without any tags).   
\end{verbatim}
\end{mdframed}

\subsubsection{Reward Model Input for Instruction Quality}
\label{app:reward_prompt}
\begin{mdframed}
\begin{verbatim}
System
A chat between a curious user and an artificial intelligence assistant. The
assistant gives helpful, detailed, and polite answers to the user's questions.

User
Passage: {{PASSAGE}}
-----
Please generate a question from the passage above.

Assistant
{{INSTRUCTION}}
\end{verbatim}
\end{mdframed}

A reward model (RM) assigns a single scalar value, i.e. a reward depending on the quality of the assistant response. Ideally, the RM learns to distinguish implicitly desirable properties of the response such as quality, factuality, helpfulness, creativity, etc.

\end{document}